%% file: main.tex
\documentclass[sigconf]{acmart}
\usepackage{booktabs} 
\usepackage{amsmath}
\usepackage{bm} 

\setcopyright{none}
\renewcommand\footnotetextcopyrightpermission[1]{} 
\usepackage[keeplastbox]{flushend}

\usepackage{hyperref}
\usepackage{booktabs}
\usepackage{graphicx}
\usepackage{subfigure}
\usepackage[normalem]{ulem}
\usepackage[labelfont=bf,font=small]{caption}
\usepackage{amsmath}

\newcommand{\figref}[1]{Fig.\ref{#1}}

\newcommand{\adnote}[1]%
 {\textcolor{blue}{\textbf{AD: #1}}}
 
\newcommand{\adrem}[1]%
 {\textcolor{green}{\textbf{CB: #1}}}
 
 \newcommand{\prg}[1]{\noindent\textbf{#1. }}

\usepackage[textfont=md,font=footnotesize]{caption}

\input{macros}

\begin{document}


\title{Learning from Richer Human Guidance: Augmenting Comparison-Based Learning with Feature Queries}
\author{Chandrayee Basu}
\affiliation{%
  \institution{UC Merced}
}
\email{basu.chandrayee@gmail.com}

\author{Mukesh Singhal}
\affiliation{%
  \institution{UC Merced}
}
\email{msinghal@ucmerced.edu}

\author{Anca D. Dragan}
\affiliation{%
  \institution{UC Berkeley}
}
\email{anca@berkeley.edu}

\input{abstract}

%
%

\keywords{reward learning, comparison-based learning, learning from human guidance, driving style}

\maketitle

\input{introduction}
 \input{math}
 \input{simulation}

\input{study}

\input{discussion}
\input{acknowledgement}

%
\bibliographystyle{abbrv}
\bibliography{featurequery}

\end{document}

%% file: macros.tex
\newcommand{\ur}{\textbf{u}}

%% file: abstract.tex
\begin{abstract}
We focus on learning the desired objective function for a robot. Although trajectory demonstrations can be very informative of the desired objective, they can also be difficult for users to provide. Answers to comparison queries, asking which of two trajectories is preferable, are much easier for users, and have emerged as an effective alternative. Unfortunately, comparisons are far less informative. 
We propose that there is much richer information that users can easily provide and that robots ought to leverage. We focus on augmenting comparisons with feature queries, and introduce a unified formalism for treating all answers as observations about the true desired reward. We derive an active query selection algorithm, and test these queries in simulation and on real users. We find that richer, feature-augmented queries can extract more information faster, leading to robots that better match user preferences in their behavior.

\end{abstract}

%% file: introduction.tex
\section{Introduction}
With robots getting better at optimizing reward functions, getting the reward function right is becoming all the more important. In many situations, specifying a reward function is challenging \cite{ng2000algorithms}. Further, specified reward functions often fail to capture some aspect that becomes important in a novel situation, leading to unintended consequences or reward hacking \cite{DBLP:journals/corr/AmodeiOSCSM16}. And most importantly, what the robot should optimize for is ultimately not up to its designer, it should be up to the end-user.

Inverse Reinforcement Learning (IRL) \cite{abbeel2004apprenticeship, ng2000algorithms, ziebart2008maximum, Ziebart:2008:NLC:1409635.1409678, ratliff2006maximum} is a natural way for robots to \emph{learn} the desired reward function. IRL collects demonstrations from a person of the desired \emph{behavior}, rather than of the desired reward, and finds parameters for the reward function that explain the demonstrated behavior. 

In this work, we focus on situations where we do not have access to demonstrations. Sometimes, demonstrations are difficult for people to provide, such as when they would need to orchestrate all of the degrees of freedom of a robot arm in a desired motion \cite{akgun2012trajectories, akgun2012keyframe}. Further, people might not know what they want until they see it. For instance, in \cite{basu2017} we showed that people think they want autonomous cars to drive like them, but in fact they want more defensive cars. Relying on their demonstrated behavior would not lead to learning the correct reward function for driving style. 

Comparison-based learning \cite{ailon2010preference, karbasi2012comparison, hullermeier2008label} has emerged as a promising alternative for learning reward in such cases \cite{akrour2012april, braziunas2006computational, daniel2014active, jain2015learning, holladayactive, christiano2017deep, dorsa2017active}. There, the robot iteratively shows users two possible trajectories (often in the same environment, for the same starting state), and asks which they prefer. It then uses the answer to update its understanding of the reward parameters.

\begin{quote}
   \emph{ We propose that robots can extract richer guidance from people when learning reward functions.}
\end{quote} 

\begin{figure}[t!]
    \centering
    \includegraphics[width=\columnwidth]{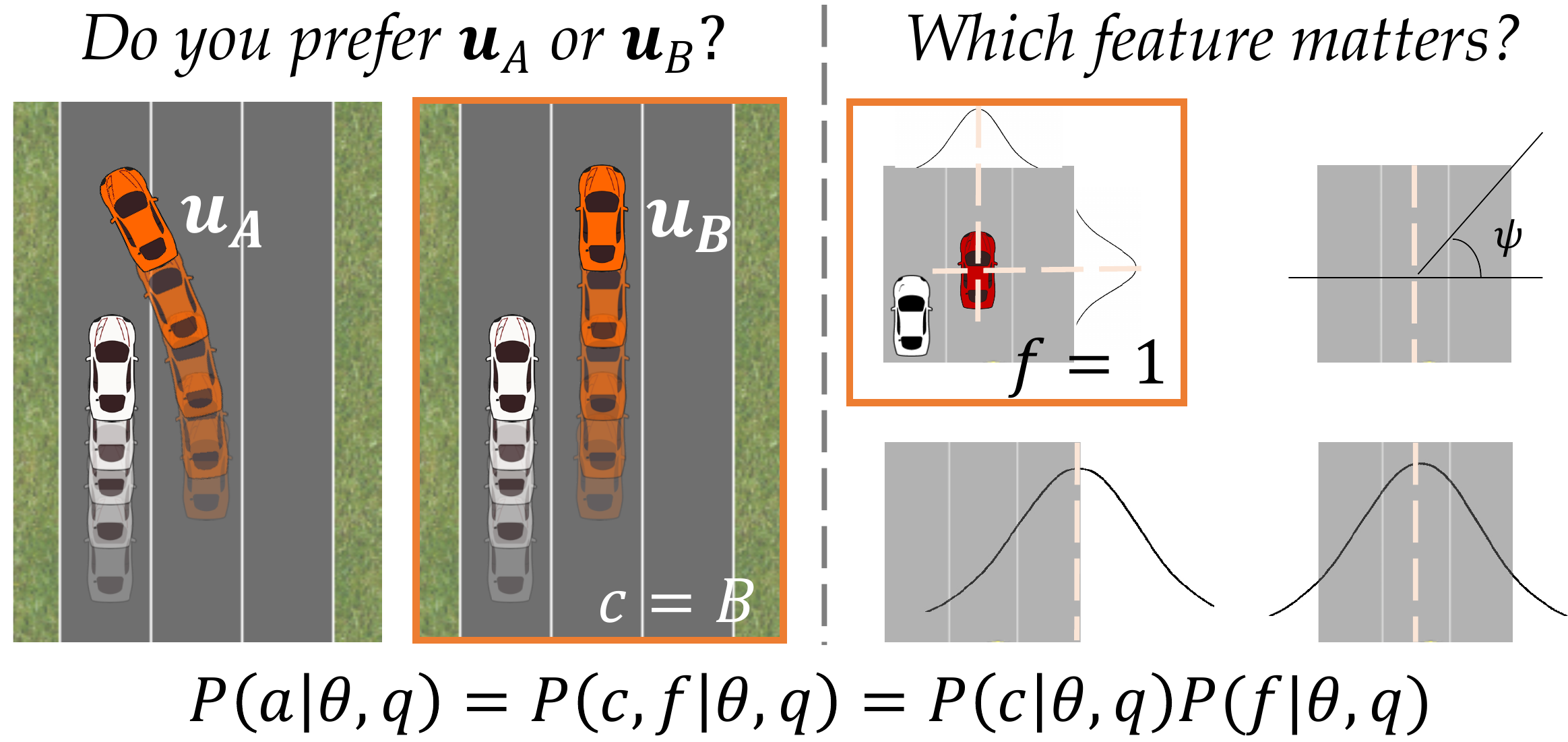}
    \caption{We introduce richer queries that augment preferences between trajectories with the features that are responsible for these preferences. We introduce a probabilistic model $P(a|\theta, q)$ of how a human might answer these queries $q$ given the desired objective function for the robot $\theta$, and use this model to actively generate rich queries that learn $\theta$.}
    \label{fig:intro}
\end{figure}

Rather than simply asking for a comparison, we introduce feature-augmented comparison queries, where the robot also asks \emph{why}: which \emph{feature} in the reward function was responsible for the preference between the two options (see \figref{fig:intro}). Feature queries have already been used in the context of learning a classifier from labels \cite{raghavan2006active}, and here we show their equivalent for learning a reward function from comparisons. 

We make the following contributions:

\prg{Rich Queries combining Comparisons and Features} We build on prior work that leveraged feature queries in the context of learning skills from demonstration \cite{cakmak2011active}  to introduce \emph{combined comparison-feature queries} for reward learning.

\prg{Learning from Rich Queries} We generalize comparison-based learning to these richer queries by treating feature answers as observations about the true reward parameters. We introduce a unifying formalism whereby the person's answers are all treated as nosily-optimal responses conditioned on the true reward, and perform Bayesian inference to estimate the reward parameters.

\prg{Active Query Selection} To speed up learning, we derive a rich query selection method that optimizes for gathering as much information as possible from each query. 

\prg{Analysis of Rich Queries} We conduct thorough experiments in simulation showing that \emph{rich queries} learn faster than \emph{comparison-only} queries, and follow-up with an in-lab study on learning driving style. We find that rich queries learn a reward that is significantly closer to the users' internal preference. This is evidenced by them preferring the robot that optimizes the reward function learned through rich queries over the robot that optimizes the reward learned through comparison-only queries. People also report a better experience with training the robot in case of rich queries.

Overall, we are excited to move away from robots expecting guidance that is difficult for real end-users to provide, and to move a step closer to robots exploiting more informative forms of guidance that people can easily provide.

%% file: math.tex
\section{Active Learning with Rich Queries}\label{sec:methods}

Our goal is to learn a reward or objective function capturing the user's preference for robot behavior, in the case in which the user can't simply provide demonstrations of what they want. We derive an algorithm for actively learning the parameters of the reward through rich queries that ask for comparisons and feature importance. 
\subsection{Problem Statement}
The robot lives in a world with state $x$, and takes actions $u$. 

We assume access to a set of human-interpretable features which can be evaluated on a tuple of state and robot action $(x,u)$. These features make a vector-valued function $\phi(x,u)$. We parametrize the robot's reward as a linear combination of these features, following the approaches in \cite{ng2000algorithms, dorsa2017active}:

\begin{equation}
r(x,u)=\theta^T\phi(x,u)
\end{equation}

Our goal is to find the parameters $\theta$ that correspond to the user's preference.

We will do so by making queries $q$ to the user, and taking their answers $a$ as \emph{observations} about the $\theta$ they want. The robot has a belief over $\theta$, which we will update at every step by:
\begin{equation}
b'(\theta)\propto b(\theta)P(a|\theta,q)
\end{equation}

In what follows we address three key remaining questions: what is a query $q$, what is our observation model for user answers $P(a|\theta,q)$, and what queries to make in order to learn efficiently.

\subsection{Queries and Observation Models for Them}
We focus on queries that present two alternative trajectories for the robot in one environment, building on prior work that has successfully learned robot rewards \cite{holladayactive,dorsa2017active}. 

Imagine learning the reward for an autonomous car, so that it drives according to the user's desired driving style. The environment is an initial state $x^0$ capturing where the road and obstacles are, along with the car's initial position, orientation, and velocity. A trajectory $\ur$ consists of a sequence of controls or actions $(u^0,..,u^T)$ for the robot. A query is thus $q=(x^0,\ur_A,\ur_B)$.

Environments can also be more complex, including the trajectories of other agents. This is particularly important, for instance, for autonomous cars, who have to share the road with other vehicles. In that case, the environment also contains an $\ur_{O}$, the trajectory for the other agents in the environment. These actions update the state of the world along with the robot's actions. 

\noindent\textbf{Comparison Only.}
Prior work has used queries of the form $q=(x^0,\ur_A,\ur_B)$.
 to ask the user "Which of the two trajectories do you prefer?". These comparison-only queries form our baseline.

Let the answer for comparison-only queries be $c\in{A,B}$: $A$ if $\ur_A$ is preferred, $B$ otherwise. Following prior work, we assume that people are noisily rational in identifying the correct trajectory:
\begin{equation}
P(c=A|\theta,q)=\frac{exp(\beta^cR(x^0,\ur_A))}{exp(\beta^cR(x^0,\ur_A))+exp(\beta^cR(x^0,\ur_B))}
\end{equation}
with $R(x^0,\ur_A)$ as the cumulative reward in the environment for trajectory $\ur_A$ and $\beta^c$ as a rationality coefficient: the higher $\beta^c$ is, the less likely is the user to make a mistake in identifying the trajectory with the higher desired reward.

\noindent\textbf{Feature Queries.} Our key idea is that we do not have to settle for comparisons only. Users have additional insight about the difference between the proposed trajectories. In particular, we propose to ask "Which feature is most responsible for the difference in your preference between these two trajectories?". 
\begin{figure*}
\includegraphics[width=\textwidth]{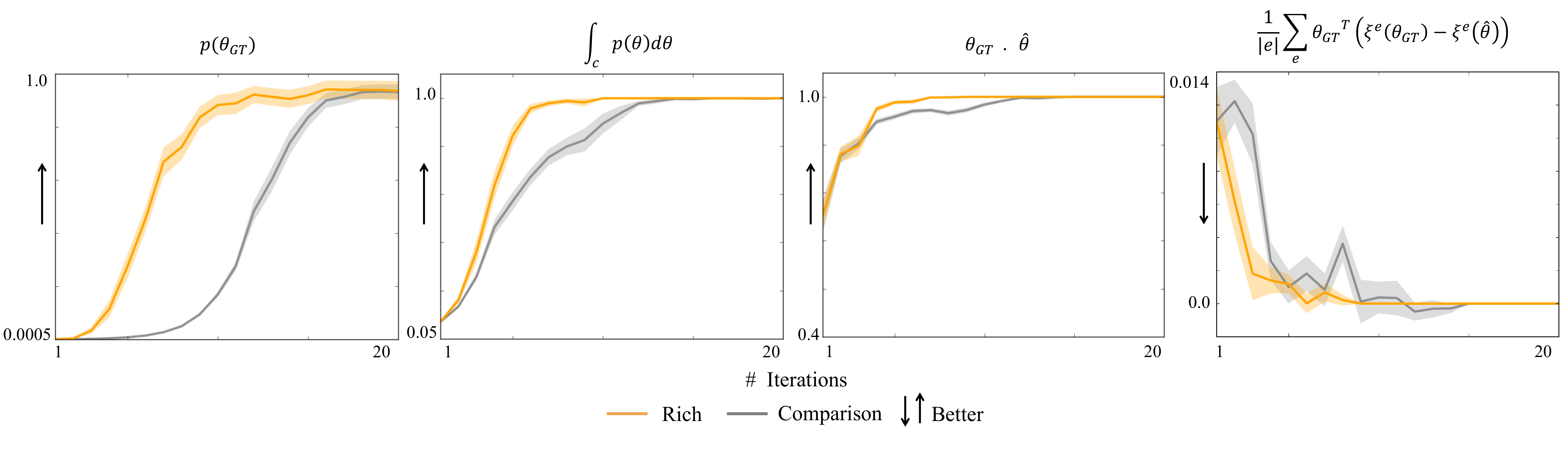}
\caption{Four dependent measures:probability of the ground truth reward function, integral over rewards similar to the ground truth, dot product between the ground truth and the learned weights, and regret, averaged across 20 true weights for each algorithm, show that compared to comparison-only queries, rich queries learn the true reward much faster, especially, when people are fully aware of their true reward.}
\label{fig:oracle}
\end{figure*}

The answer is a feature ID $f$ out of the set of $F$ features that the robot is using in its reward. We introduce an observation model for such queries, based on the assumption that people will noisily identify the feature $f$ which, when combined with how important it is (its weight in $\theta$), best accounts for the difference in reward between the two trajectories:
\begin{equation}
    P(f|\theta,q)=\frac{exp(\beta^f \theta_f \cdot |\Phi_f(x^0,\ur_A)-\Phi_f(x^0,\ur_B)|)}{\sum_i exp(\beta^f \theta_i \cdot |\Phi_i(x^0,\ur_A)-\Phi_i(x^0,\ur_B)|)}
\end{equation}
with $\beta^f$ a rationality coefficient for these types of queries, $\Phi_f$ the value of $f^{th}$ feature accumulated across the trajectory, and $\theta_f$ its weight in $\theta$.

\noindent\textbf{Rich Queries.} We propose rich queries that combine comparison and feature queries. From the answer to the comparison, the robot will know which trajectory the user prefers. From the answer to the feature query, the robot will know more about \emph{why} that is.

The answer is a tuple $a=(c,f)$. Each part serves as an observation about $\theta$:
\begin{equation}
   p(a|\theta,q)= P(c,f|\theta,q)=P(c|\theta,q)P(f|\theta,q)
\end{equation}
We model the answers as conditionally independent given the reward -- this is a design choice, and it means that people don't have to get the comparison correct to tell us the main feature that matters in the comparison. Alternately, we could condition the feature on the comparison answer, and model a feature as only probable when the sign of the weighted feature difference, not just the absolute value, is consistent with the weight vector. 

\subsection{Active Query Selection}
So far we have defined our queries, and how to update the robot's belief given the answer to a query. Now, we turn to which queries to make, i.e. how the robot should select queries.

The simplest option for query selection is to draw queries at random from a set. However, we can speed up learning by \emph{actively} selecting queries. 

At every step, we would ideally select the query that will remove as many $\theta$s from the hypothesis space as possible. There are two challenges with this. 

The first challenge is that we do not eliminate hypotheses, rather update a belief over them. We thus want to maximize some measure of the change in probability distribution over $\theta$. Following \cite{dorsa2017active}, we use volume removed: a query $q$ with an answer $a$ removes the following volume:
\begin{equation}
 V(q,a)=\sum_{\theta}P(\theta)-P(\theta)P(a|\theta,q) = \mathbb{E}_{\theta}[1-P(a|\theta,q)]
\end{equation}
with our $P(a|\theta,q)$ from our definition above.

The second challenge is that we don't know what answer we will get. However, our current belief induces a probability distribution over answers, so we can choose the query that \emph{in expectation} removes the most volume:
\begin{equation}
    \max_q \mathbb{E}_{\theta}\sum_{a} P(a|\theta,q) V(q,a)
\end{equation}
where the sum over $a$ is over all tuples $(c,f)$. 

Note that other measures are also possible, including information gain (reduction in Shannon entropy) \cite{settles2010active}.  

\subsection{Allowing "I don't know"}
Since feature queries are more complex than their comparison counterpart, we also experiment with allowing users to say "I don't know". In that case, we skip the update based on the feature query, and only use the answer to the comparison query for update, i.e. $P(c|\theta,q)$.

More interestingly, we inform the query selection criterion about this option, so that it does not choose queries that will lead to no update from the feature response. 

To achieve this, we model \emph{when} a user might say that they don't know as their answer to the feature query. We model this "I don't know answer" as occurring whenever the probability distribution over their answers is too close to uniform, i.e. 
\begin{equation} \label{eq:12}
    P(skip|\theta,q) = \begin{cases}
    1 \quad \frac{1}{d} - \epsilon \leq p(\mathbf{f}|\mathbf{\theta},\varphi) \leq \frac{1}{d} + \epsilon\ \forall f\in\{1..d\}\\
    0 \quad otherwise
    \end{cases}
\end{equation}
with $d$ the number of features and $\epsilon$ capturing how similar the contributions of the features need to be with respect to each other for the user to decide to skip the answer.

If the person does skip, then the volume removed is 
\begin{equation}
    V^{skip}(q,c)=\mathbb{E}_{\theta}[1-P(c|\theta,q)]
\end{equation}

Then the robot selects the next query by optimizing
\begin{align}
\max_q \ &\nonumber \mathbb{E}_{\theta} P(skip|\theta,q)\cdot \sum_c P(c|\theta,q)V^{skip}(q,c) +\\ &(1-P(skip|\theta,q))\cdot \sum_a P(a|\theta,q)V(q,a) 
\end{align}

%% file: simulation.tex
\section{Hypothesis}
\emph{We hypothesize that using rich queries (i.e. our queries that seek feature clarification) leads to higher accuracy of the learned reward, given the same budget of queries.} We test this hypothesis in simulation, and in a user study.

\begin{figure*}
\includegraphics[width=\textwidth]{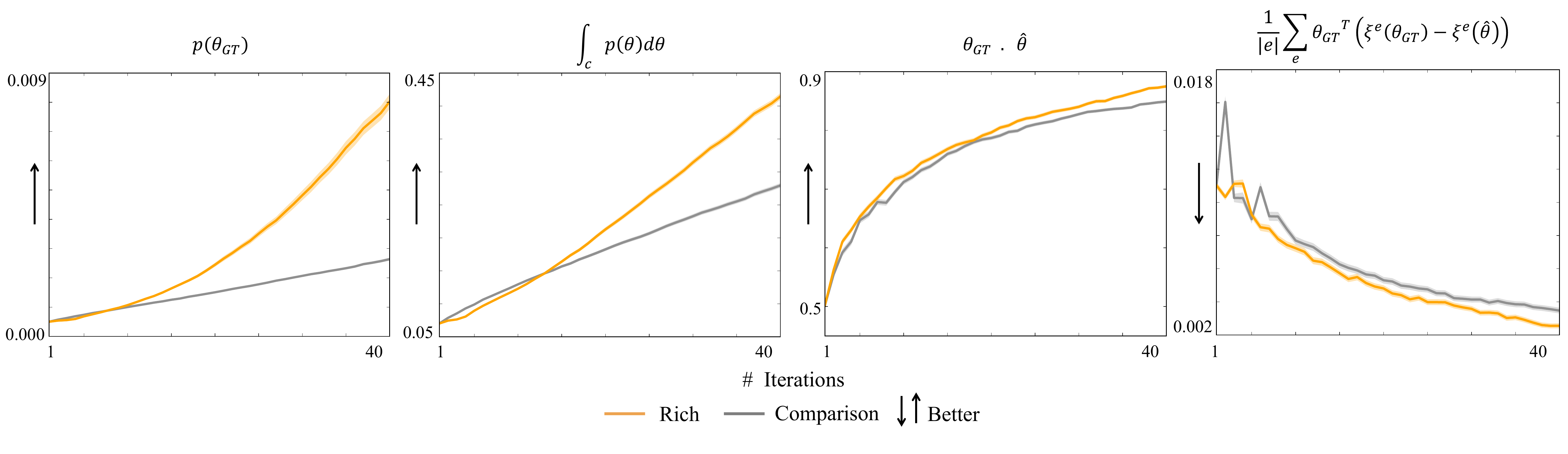}
\caption{Four dependent measures: probability of the ground truth reward function, integral over rewards similar to the ground truth, dot product between the ground truth and the learned weights, and regret, averaged across 20 ground truth weights. The results show that even when people provide noisy responses to feature queries, the robot still learns better compared to comparison-only queries.}
\label{fig:noisy}
\end{figure*}

\section{Experiments in Simulation}
We start with experiments that simulate user responses based on some known true weights $\theta_{GT}$.

\subsection{Experiment Design}

\noindent\textbf{Manipulated Factors.}
We manipulated the type of queries we used for learning, with two levels: \emph{comparison} only queries vs. \emph{rich} queries that ask for a comparison, along with the feature that is important in the difference.

We also manipulated the number of queries that each resulting algorithm gets to make, from 0 to 40. 

Finally, we repeated the experiment for 20 possible ground truth weights.

\noindent\textbf{Dependent Measures.} We evaluated the accuracy of the learned probability distribution over reward parameters $\theta$ in four ways:
\begin{itemize}
\item The learned probability of the true weights: $$P(\theta_{GT})$$ 
The higher the probability assigned to the ground truth, the more accurate the learned distribution is.
\item The integral of the probability of all "close" weights, i.e. all weights that have high dot product with the ground truth weight:
$$\int_C P(\theta)d\theta$$
with $C=\{\theta|\theta \cdot \theta_{GT}>0.9\}$. This is a more robust version of the first measure, where we capture not only the probability of the ground truth, but also what the probability distribution does around the ground truth. 
\item The dot product between the ground truth weights and the learned weights:
$$\theta_{GT} \cdot \hat{\theta}$$ 
with $\hat{\theta}=\arg\max_{\theta}P(\theta)$. Rather than focusing on just the ground truth, this measure gets at how good the weights that we learned, $\hat{\theta}$, actually are.
\item The regret resulting from converging to a reward function different from the true reward function measured across different test environments:
$$\frac{1}{|e|}\sum_e \theta_{GT}^T (\xi^e(\hat{\theta}) - \xi^e(\theta_{GT})) $$
with $\xi^e(\hat{\theta})$ denoting the optimal trajectory in environment $e$ for the reward function $\hat{\theta}$ and $\xi^e(\theta_{GT})$ denoting the optimal trajectory for the true reward function. In other words, we measure the difference in true reward between the \emph{actually} optimal trajectory, and the \emph{learned} optimal trajectory. This captures what happens when we go out into the world and optimize the reward that we learned, i.e. to what extent do we match the user's actual preference. 

\end{itemize}

\subsection{Problem Domain}
We focus on learning \emph{driving} preferences. 

\noindent\textbf{Features for the Reward Function.} Our $\theta$s weigh seven features for driving (see Fig. \ref{fig:intro}):
\begin{itemize}
\item $f_1$ exponentially decreases as the distance to the center of the lane increases
\item $f_2$ exponentially decreases as the distance to the edge of the road increases
\item $f_3$ computes alignment between the car's heading and the road
\item $f_4$ computes distance to other cars, as in \cite{dorsa2017active}
\item $f_5$ increases with the speed of the car
\item $f_6$ incentivizes a preference to be in the right lane
\item $f_7$ disincentivizes reverse motion
\end{itemize}

\noindent\textbf{Queries.}
Queries consist of an environment and two trajectories for that environment shown in a simulator (see Fig. \ref{fig:intro}). The environment consists of an initial state for the user's car, along with a trajectory for another traffic car with which  the user's car shares the road. Our query trajectories for the user's car are always optimal with respect to \emph{some} reward, such that users are essentially comparing reward functions. We pre-computed a query pool consisting of 7000 queries generated from a combination of 19 "plausible" reward functions and 40 environments. We rejection-sampled plausible reward functions by eliminating rewards that, for instance, incentivise the car to crash.

\subsection{Oracle Users}
\noindent\textbf{The Users We Simulate.} 
In our first experiment, we tested our hypothesis for the case that users are perfect, meaning their answers are noise-free. The user becomes an oracle, who is fully aware of the ground truth reward $\theta_{GT}$ and uses it to answer queries, returning $I^{*}= argmax  P(I|\theta,\varphi,\beta^c)$ for the comparison and $i^{*}=argmax P(i|\theta,\varphi,\beta^f)$ for the most influencing feature. Equivalently, this oracle user has $\beta^c=\beta^f=\infty$.

We thus \emph{simulated} oracle answers based on $\theta_{GT}$, setting the simulated noise $\beta^c_s=\beta^f_s=\infty$. We refer to the simulation noise parameters as $\beta^{c}_{s}$ and $\beta^f_s$, where {s} stands for \emph{simulation}. We distinguish these parameters from model assumption of user noise, which we denote by $\beta^{c}_{m}$ and $\beta^f_m$ ({m} stands for \emph{model}).

\noindent\textbf{Our Model of the Users.}
First we used an accurate model of the oracle behavior: the learning algorithm models users as perfect as well, with the modeled noise parameters $\beta^c_m=\beta^f_m=\infty$.

\noindent\textbf{Analysis.}
For each ground truth reward function and the number of queries allowed, a query in each learning method produces a probability distribution over reward parameters, which we evaluate according to our dependent measures.
We analyzed the effect of the query type on each measure using a repeated measures ANOVA. Here we included the number of queries to help explain the difference between both algorithms' performance with very few queries and with many queries. We included an identifier for the ground truth weight as a random effect. 

The results support our hypothesis. Rich queries lead to significantly higher probability being assigned to the ground truth reward function ($F(1,778)=374.02$, $p<.0001$), significantly higher integral over rewards similar to the ground truth ($F(1,778)=33.60$, $p<.0001$)), significantly higher dot product between the ground truth and the learned weights ($F(1,778)=4.89$, $p=.02$), and a significantly lower regret when used to optimize behavior ($F(1,778)=10.14$, $p<.01$).

\figref{fig:oracle} summarizes these results. 

\begin{figure}
\includegraphics[width=0.5\textwidth]{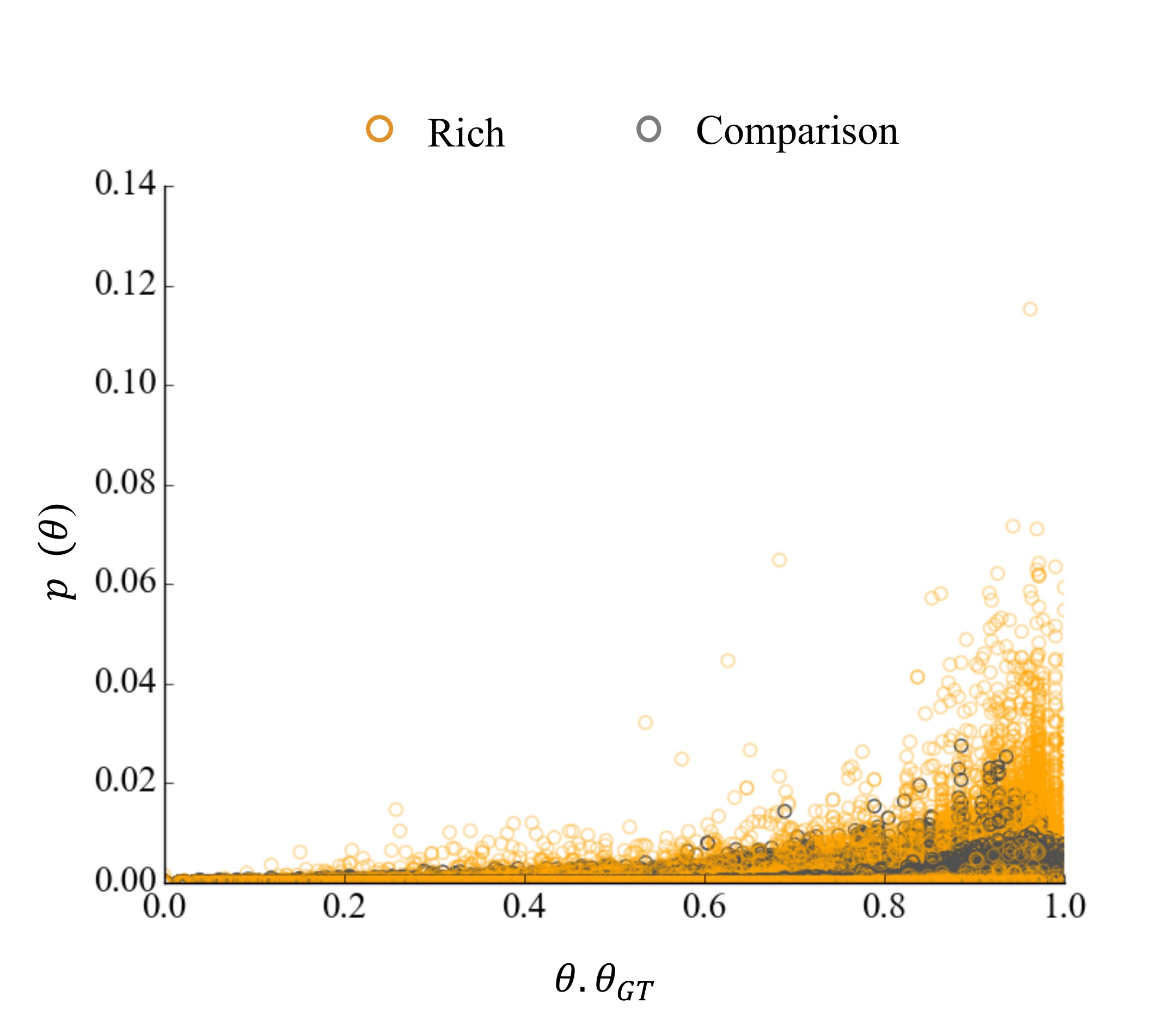}
\caption{A scatter plot of the probability of all the weights in the data set by their dot product with a true reward shows that our algorithm converges much faster than comparison-only queries. Here $\beta^{f}_{s}$ = $\beta^{f}_{m}$ = 2.5, $\beta^{c1}_{s}$ = $\beta^{c1}_{m}$ = 5. and $\beta^{c2}_{s}$ =  $\beta^{c2}_{m}$ = 2.}
\label{fig:noisy2}
\end{figure}

\subsection{Realistic, Noisy Users}\label{Noisy}
\noindent\textbf{The Users We Simulate.}
Real people are imperfect. They will not follow the $\beta^c=\beta^f=\infty$ assumption. 
We thus ran a pilot user study with a few real users to estimate the realistic values for $\beta^c$ and $\beta^f$. We estimated the users'  true reward parameters, then generated queries in different environments and asked them to provide responses to the queries. We ran a maximum likelihood estimation for the $\beta$s given their data:
\begin{equation}\label{eq:13}
    \beta^{c*} = argmax\sum_{i = 1}^N\log\frac{1}{1 + exp(-I_{i} \beta \theta^{*T}\Phi_i^*)}
\end{equation}

\begin{equation}\label{eq:14}
    \beta^{f*} = argmax\sum_{i = 1}^N\log\frac{exp(|\beta^{f} \theta_f^* \phi_{if})|}{(1 + exp(-I_{i} \beta^{c} \theta^{*T}\Phi_i))\sum_j{exp(|\beta^{f} \theta_j^*\phi_{ij}|)}}
\end{equation}
Here $|N|$ denotes total number of queries made and $\theta_f^*\phi_{if}$ is the true reward contribution of the most influencing feature for that query.
We decoupled the comparison query function and the feature query function in equation \ref{eq:14} and performed a search for $\beta^{c}$ over several thousand values of $\beta$ and then used this $\beta^{c}$ to run a similar search for $\beta^{f}$. We also computed $\beta^{c*}$ separately for rich queries and comparison queries. To our surprise, we found that ${\beta}^{c}$ is much higher for our algorithm than for the comparison-only method: users were more accurate in comparison responses when the comparison queries are accompanied by feature queries. We found the average ${\beta}^{c}$ to be 1.6 and 5.65 for the baseline and for our algorithm respectively. Objectively, this might be due to the difference in the nature of the queries selected by the two algorithms. Our algorithm selects queries with low norms that differ along only very few features, in order to make the feature answer more useful. This might lead to comparisons that are easier to make. Subjectively, too, most users seemed to think they were better at answering comparisons when they had feature queries too (the average on a 7-point Likert scale for the question "I thought answering feature queries improved my ability to compare between the two trajectories" was 6.0). 

\begin{figure*}[t!]
    \centering
    \includegraphics[width = \textwidth]{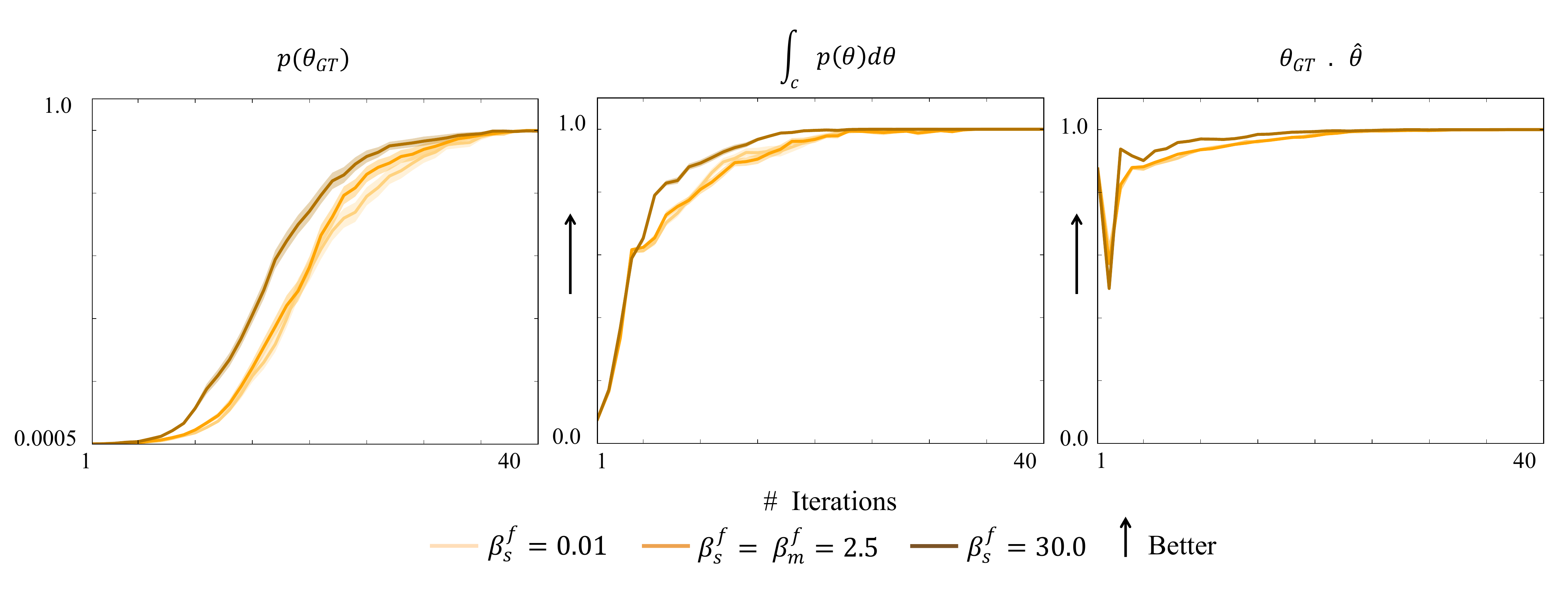}
    \caption{As users become more or less noisy in answering the feature query than we model them to be (captured by $\beta^f_s$), our algorithm's performance ranges between the performance of learning from comparisons-only queries, and that of learning from oracle users.} 
    \label{fig:betavariation}
\end{figure*}

\begin{figure*}[t!]
    \centering
    \includegraphics[width = \textwidth]{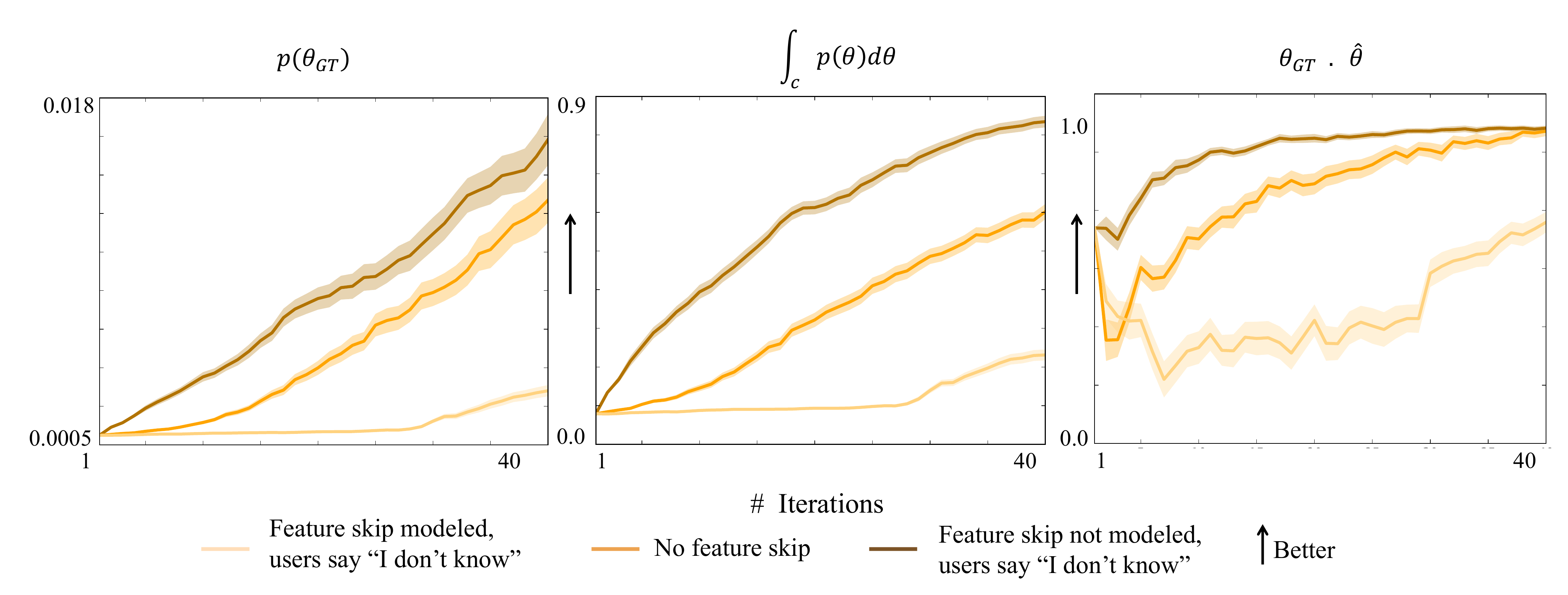}
    \caption{Graphs showing three dependent measures for the case when users are allowed to skip feature queries and this user behavior is incorporated in the model. The dependent measures are for a single true reward function averaged over 100 repetitions. Allowing users to skip the feature queries they are unsure of improves the performance of our algorithm. But interestingly, incorporating this behavior into the model hurts the performance.}
    \label{fig:fskipiteration}
\end{figure*}

We simulated imperfect real users with these estimated values of $\beta^c_s$ and $\beta^f_s$. Following the results of the pilot we considered two different noise parameters for comparison answers: $\beta^{c1}_s$ for the comparison-only method and $\beta^{c2}_s$ for rich queries. Note that ${\beta}^{f}$ = 0 is equivalent to the baseline, where we get no additional information because here the users are totally random in answering the feature queries and ${\beta}^{f}$ = $\infty$ is equivalent to oracle feature picking. 

\noindent\textbf{Our Model of the Users.}
In this experiment, we first assumed that the models have accurate knowledge of the user behavior and set $\beta^{c1}_m=\beta^{c1}_s$ for the comparison-only algorithm and $\beta^{c2}_m=\beta^{c2}_s$ and $\beta^f_m=\beta^f_s$ for our algorithm. This scenario replicates what would happen when the learning algorithm has a good model of what noise level to expect from users. 

\noindent\textbf{Analysis.}
This time, since we simulate users to be noisy, we ran each simulated condition, which is a combination of ground truth weight and query type, 100 different times with different random seeds for each of the 20 ground truth weights. We ran these simulations in parallel using PyWren \cite{pywren}. 

We repeated the analysis from before, and found support for our hypothesis despite the fact that we no longer learn from perfect user answers. Rich queries lead to significantly higher probability assigned to the ground truth reward, significantly higher integral over similar rewards to the ground truth, significantly higher dot product of the learned reward with the true reward, and significantly lower regret, all with $F(1,1e+5)>335$ and $p<.0001$ throughout. The scatter plot between the probability distribution over the reward space and their dot product with the true reward, in fig. \ref{fig:noisy2} shows a comparison between the convergence of the two algorithms. Our algorithm with feature queries converges faster to a reward function closer to the true reward.

This is not too surprising: even with noisy users, that the added information about features should still help. \figref{fig:noisy} shows the results.

\subsection{Users with Different Noise Level from the Model}
\noindent\textbf{Simulated Users and Model.}
In some situations, it will indeed be possible to get a good estimate of the user noise level. Here, we explore what happens when this is not the case, and the users are either much worse or much better at answering feature queries than we expect. We set $\beta^f_m$ as before, but vary $\beta^f_s$, namely the simulated user noise. Throughout these experiments, we keep $\beta^{c}=\infty$ to avoid additional noise introduced through noisy comparison response.

\noindent\textbf{Analysis.}
Not surprisingly, as $\beta^f_s$ ranges from 0 to $\infty$, our algorithm's performance improves, interpolating between the comparison-only performance and the oracle performance. Note that $\beta^{f}_m = 2.5$, used in the noisy model, is less than the estimated optimal value of 3.8 from pilot studies. Fig. \ref{fig:betavariation} shows that this is quite a conservative value. The algorithm performance does not decrease significantly with lower values of $\beta^{f}$. On the other hand, if people are more accurate about their true reward functions, feature queries can lead to a much faster learning. 
\begin{figure}[t!]
    \centering
    \includegraphics[width = 0.5 \textwidth]{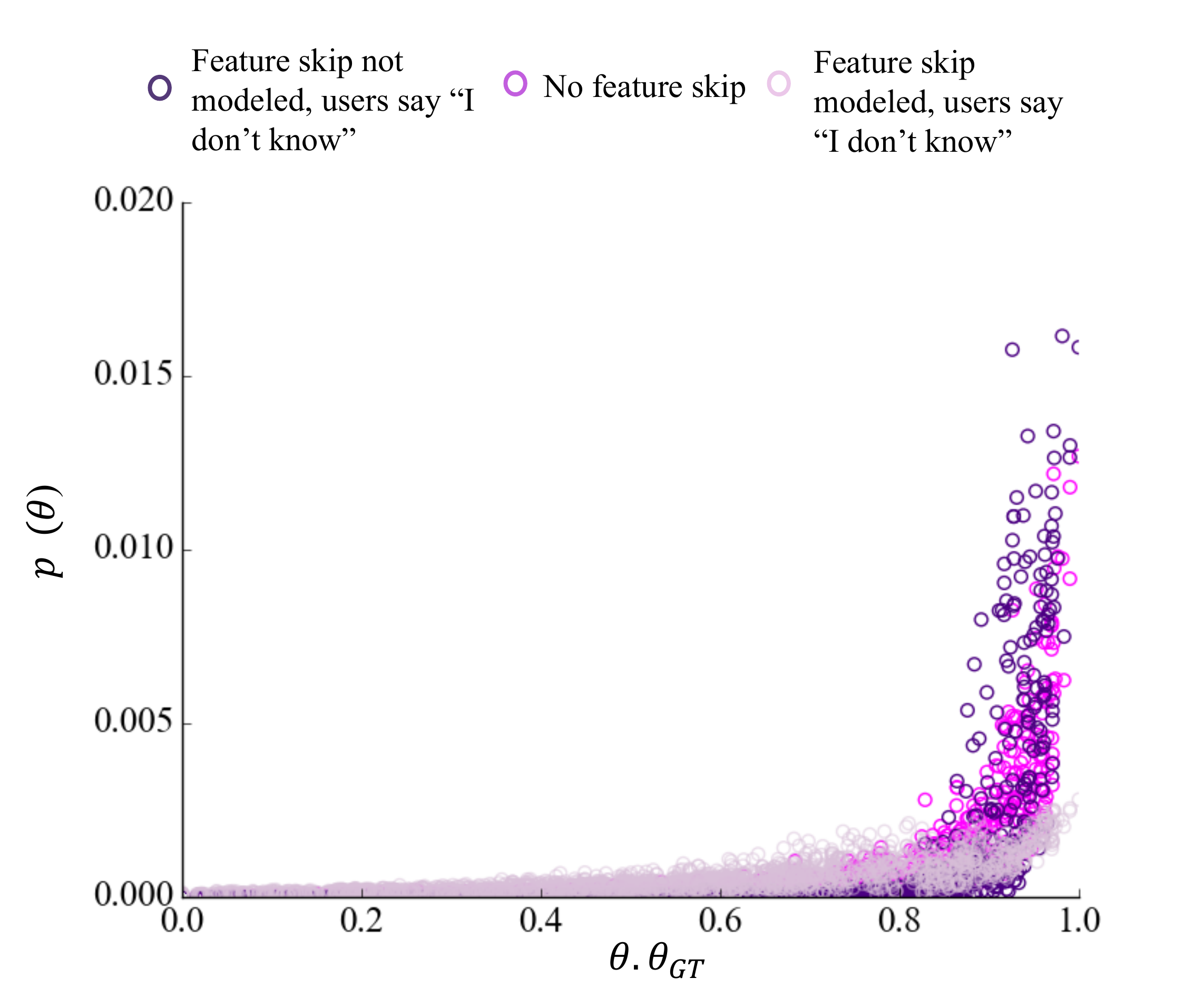}
    \caption{Scatter plot of the probability of all reward functions by their dot product with the true reward function at convergence. This graph suggests that allowing users to skip the feature queries they are unsure of leads to faster convergence. But incorporating this human behavior into the model hurts convergence.}
    \label{fig:fskipscatter}
\end{figure}

\begin{figure*}[t!]
    \centering
    \includegraphics[width = \textwidth]{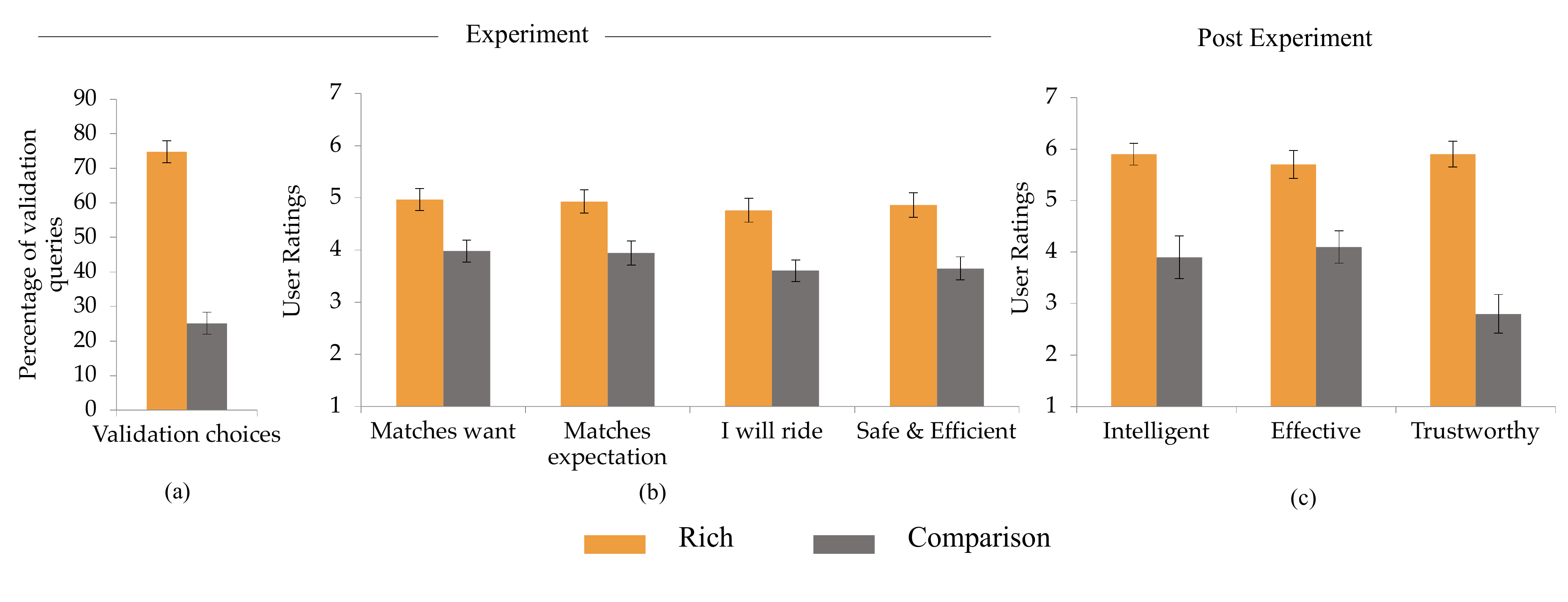}
    \caption{(a) During the validation phase most participants preferred the trajectories optimal for $\theta^*_{rich}$. (b) Overall trajectories learned with feature queries matched what participants' desired and expected driving behavior and also appeared to be safer and efficient. (c) Post experiment, most participants agreed that the robot with feature queries was more intelligent, effective and trustworthy than the one that just made preference query. }
    \label{fig:userratings}
\end{figure*}

\subsection{Users that Say "I don't know."}
\noindent\textbf{The Users we Simulate.}
We now turn to studying the utility of allowing users to say "I don't know", i.e. to skip the feature part of the rich queries. We recover from a pilot study analogous to the one on the noise parameters a value for $\epsilon$ at 0.066. We used this value of   $\epsilon_{s}$ to simulate no response to the feature query. We ran these simulations assuming noisy responses to comparison and feature queries as before. 

\noindent\textbf{Our Model of the Users.}
Here we switch to the learning algorithm that models users as saying "I don't know" within some $\epsilon_m$ (see eq. \ref{eq:12}), and set $\epsilon_m=\epsilon_s$.

\noindent\textbf{Analysis.}
The results show that it is wise to avoid responding to feature queries when highly unsure, under the current assumption of noise ($\beta^f_m$ = 2.5), matching our expectations. Our algorithm performs better when we allow people to say "I don't know" compared to when we do not allow them to. For a given true weight, we observed that after 40 iterations, the probability of the true weight improves from 0.012 to 0.016 as we allow people to skip feature queries. Likewise, allowing skipping leads to learning weights that are much closer to the true weight (probability integral of close $\theta$s = 0.8) than when people respond to every query (probability integral of close $\theta$s = 0.6). Fig. \ref{fig:fskipiteration} shows these results. This result is also apparent from the scatter plot in Fig. \ref{fig:fskipscatter} that shows a faster convergence when people are allowed to say "I don't know". 

However, to our surprise, if we include the assumption that people might say "I don't know" in our model, the performance of the algorithm suffers. For the same true weight as above, we found the probability of this true weight and the probability integral of close $\theta$s to be 0.003 and 0.23 respectively, compared to 0.016 and 0.9 when feature skipping is not modeled. Furthermore, the scatter plot between probability distribution over weight space and dot product with true weights hardly shows any sign of convergence. This performance is comparable with the case when no feature query is made. Intuitively, having a better user model should lead to higher performance, but it seems like at least the volume removed metric (which optimizes purely for information) interacts poorly with this user model and makes active query selection less useful.

%% file: study.tex
\section{User Study} \label{studies}
\subsection{Experiment Design}

Next, we tested our hypothesis with actual users.

\noindent\textbf{Study Protocol Overview.} Each user answered 20 queries of each type: comparison-only and rich. For each method, we used the 20 answers to learn a reward function: two reward functions per participant, that we then compared in order to test which method learned a more accurate reward. We then used 10 test environments for validation. We asked the participants, for each test environment, to choose between a pair of trajectories (one optimal with respect to the reward learned through comparisons only, one optimal with respect to the reward learned through rich queries). We also asked them to rate each of the two trajectories in each environment on Likert scales for the statements in Table \ref{tab:likert}. Finally, at the end of the study, the participants answered questions about the experience of interacting with rich queries (Table \ref{tab:post experiment}). 

\noindent\textbf{Manipulated Factors.} We manipulated the query type. Informed by the results of our simulation, we also allowed participants to skip queries if they were unsure of the answers. 

\noindent\textbf{Participants and Allocation Method.} We recruited 10 participants consisting of a mix of undergraduate and graduate students, and used a within-subjects allocation so that participants could actually compare the outcome of the two learning methods. All the participants had 1+ years experience in driving and were comfortable with feature queries even with limited technical background. 

\noindent\textbf{Dependent Measures.} Since with real people we no longer have access to the ground truth reward (it is internal to them), we had to settle on a different way to measure the quality of the learned reward. We thus had participants compare the learned rewards from each method by exposing them to two trajectories that optimize each reward in 10 test environments. 

For each environment and trajectory, we asked participants 4 Likert-scale questions, capturing their preference for one or the other (see Table \ref{tab:likert}). We also asked them, for each environment, to choose the trajectory they prefer. These questions helped us understand if the users considered any of the two methods to be particularly effective or both methods to be ineffective.

{\renewcommand{\arraystretch}{1.01}
\begin{table}[h!]
  \caption{Measures for each trajectory in each test scenario.}
  \label{tab:likert}
  \begin{tabular}{l}
    \multicolumn{1}{c}{\textbf{Validation Likert Questions}}  \\
    \hline
    \multicolumn{1}{p{8cm}}{Q1: Trajectory X matches what I want the car to do.} \\
    \multicolumn{1}{p{8cm}}{Q2: Trajectory X is what I would expect the car to do.} \\
    \multicolumn{1}{p{8cm}}{Q3: I would like to ride in the car using trajectory X.} \\
    \multicolumn{1}{p{8cm}}{Q4: I think trajectory A is the right combination of safety and efficiency.} \\
    
    \hline
  \end{tabular}
\end{table}}

\subsection{Analysis}
We analyzed participants' ratings of the trajectories produced by each learning method using a repeated-measures ANOVA for each item. We found significant effects of the query type across the board: rich queries led to significantly higher evaluation of whether the car matched what the user wanted ($F(1,175)=11.52$, $p=<.001$), whether it matched what they expected ($F(1,175)=10.00$, $p=<.01$), whether they would like to ride it ($F(1,175)=14.41$, $p=<.001$), and whether it was the right combination of safety and efficiency ($F(1,175)=15.05$, $p=<.001$). The trajectory corresponding to the weight learned by the rich queries was chosen 74\% of the time. 

Participants found the robot that makes rich queries more intelligent, effective and trustworthy. Fig. \ref{fig:userratings} shows the results of the user study experiment and their responses to the post-study Likert Scale questions.

We asked participants to rate their experience with the feature queries on a 7-point Likert scale in the following form:
 {\renewcommand{\arraystretch}{1.01}
\begin{table}[h!]
  \caption{Feedback on experience with feature queries.}
  \label{tab:post experiment}
  \begin{tabular}{l}
    \multicolumn{1}{c}{\textbf{Post-study Likert Questions}}  \\
    \hline
    \multicolumn{1}{p{8cm}}{Q1: I thought answering feature queries was useful.} \\
     \multicolumn{1}{p{8cm}}{Q2: The method where the car uses feature queries was annoying.} \\
    \multicolumn{1}{p{8cm}}{Q3: I thought answering feature queries gave more transparency into the working of the car.} \\
    \multicolumn{1}{p{8cm}}{Q4: I thought answering feature queries improved my ability to compare between the two trajectories.} \\
    \multicolumn{1}{p{8cm}}{Q5: I liked when I did not have to answer feature queries.} \\
    \hline
  \end{tabular}
\end{table}}

Most participants considered feature queries to be extremely useful (average rating 5.8 and that answering feature queries helped them to understand how the car worked and thus improve their comparison capabilities (average ratings 5.9 and 6.3 respectively). They all agreed that they would help the car to learn their preferences better by answering feature queries, even if it was extra work (average rating for Q5 in Table \ref{tab:post experiment}). Overall the participants reported positive experience with feature queries and gave an average rating of 6.0 over all the experience-related statements.

%% file: discussion.tex
\section{Discussion}
\noindent\textbf{Summary.}
The key idea of this paper is that we robots can extract richer information from people about their preferences if they make the right kinds of queries. 
We introduced feature queries as a way to augment comparison-only queries and get richer guidance from users when learning reward functions. We did an in-depth analysis in simulation, emulating perfect and noisy responses. We found that the richer queries consistently outperform comparison-only queries in being able to extract the correct reward faster. We then did an in-lab user study where participants interacted with each learning method, and again found that rich queries led to better outcomes within the same number of iterations.

\noindent\textbf{Limitations and Future Work.}
While rich queries are really helpful, they assume that the features of the reward function are interpretable, and that they can be explained to end-users. This was the case in our application, but it will not always hold. There has been a lot of progress in learning features from sensorimotor data directly via deep learning, and even though there is much excitement about being able to interpret or visualize these features, this is still work in progress.

Nonetheless, we are excited to have taken one more step in a research agenda of not letting algorithms dictate what guidance robots get from people. Rather, we adjust algorithms to  leverage  informative guidance that people can easily provide beyond just comparisons. 

%% file: acknowledgement.tex
\section{Acknowledgement}
This work is supported by AFOSR FA9550-17-1-0308, NSF 1652083, and Berkeley Deep Drive. We would like to thank Narayanan Sundaram and Dylan Hadfield-Menell for valuable discussions.

%% file: main.bbl
\begin{thebibliography}{10}

\bibitem{pywren}
pywren.
\newblock \url{http://pywren.io/}.
\newblock Accessed: 2017-12-24.

\bibitem{abbeel2004apprenticeship}
P.~Abbeel and A.~Y. Ng.
\newblock Apprenticeship learning via inverse reinforcement learning.
\newblock In {\em Proceedings of the twenty-first international conference on
  Machine learning}, page~1. ACM, 2004.

\bibitem{ailon2010preference}
N.~Ailon and M.~Mohri.
\newblock Preference-based learning to rank.
\newblock {\em Machine Learning}, 80(2):189--211, 2010.

\bibitem{akgun2012keyframe}
B.~Akgun, M.~Cakmak, K.~Jiang, and A.~L. Thomaz.
\newblock Keyframe-based learning from demonstration.
\newblock {\em International Journal of Social Robotics}, 4(4):343--355, 2012.

\bibitem{akgun2012trajectories}
B.~Akgun, M.~Cakmak, J.~W. Yoo, and A.~L. Thomaz.
\newblock Trajectories and keyframes for kinesthetic teaching: A human-robot
  interaction perspective.
\newblock In {\em Proceedings of the seventh annual ACM/IEEE international
  conference on Human-Robot Interaction}, pages 391--398. ACM, 2012.

\bibitem{akrour2012april}
R.~Akrour, M.~Schoenauer, and M.~Sebag.
\newblock April: Active preference learning-based reinforcement learning.
\newblock In {\em Joint European Conference on Machine Learning and Knowledge
  Discovery in Databases}, pages 116--131. Springer, 2012.

\bibitem{DBLP:journals/corr/AmodeiOSCSM16}
D.~Amodei, C.~Olah, J.~Steinhardt, P.~Christiano, J.~Schulman, and
  D.~Man{\'{e}}.
\newblock Concrete problems in {AI} safety.
\newblock {\em CoRR}, abs/1606.06565, 2016.

\bibitem{basu2017}
C.~Basu, Q.~Yang, D.~Hungerman, M.~Singhal, and A.~D. Dragan.
\newblock Do you want your autonomous car to drive like you?
\newblock In {\em Proceedings of the 2017 ACM/IEEE International Conference on
  Human-Robot Interaction}, pages 417--425. ACM, 2017.

\bibitem{braziunas2006computational}
D.~Braziunas.
\newblock Computational approaches to preference elicitation.
\newblock {\em Department of Computer Science, University of Toronto, Tech.
  Rep}, 2006.

\bibitem{cakmak2011active}
M.~Cakmak and A.~L. Thomaz.
\newblock Active learning with mixed query types in learning from
  demonstration.
\newblock In {\em in Proc. of the ICML Workshop on New Developments in
  Imitation Learning}. Citeseer, 2011.

\bibitem{christiano2017deep}
P.~Christiano, J.~Leike, T.~B. Brown, M.~Martic, S.~Legg, and D.~Amodei.
\newblock Deep reinforcement learning from human preferences.
\newblock {\em arXiv preprint arXiv:1706.03741}, 2017.

\bibitem{daniel2014active}
C.~Daniel, M.~Viering, J.~Metz, O.~Kroemer, and J.~Peters.
\newblock Active reward learning.
\newblock In {\em Robotics: Science and Systems}, 2014.

\bibitem{dorsa2017active}
A.~D.~D. Dorsa~Sadigh, S.~Sastry, and S.~A. Seshia.
\newblock Active preference-based learning of reward functions.
\newblock In {\em Robotics: Science and Systems (RSS)}, 2017.

\bibitem{holladayactive}
R.~Holladay, S.~Javdani, A.~Dragan, and S.~Srinivasa.
\newblock Active comparison based learning incorporating user uncertainty and
  noise.
\newblock In {\em RSS Workshop on Model Learning for Human-Robot
  Communication}, 2016.

\bibitem{hullermeier2008label}
E.~H{\"u}llermeier, J.~F{\"u}rnkranz, W.~Cheng, and K.~Brinker.
\newblock Label ranking by learning pairwise preferences.
\newblock {\em Artificial Intelligence}, 172(16):1897--1916, 2008.

\bibitem{jain2015learning}
A.~Jain, S.~Sharma, T.~Joachims, and A.~Saxena.
\newblock Learning preferences for manipulation tasks from online coactive
  feedback.
\newblock {\em The International Journal of Robotics Research},
  34(10):1296--1313, 2015.

\bibitem{karbasi2012comparison}
A.~Karbasi, S.~Ioannidis, et~al.
\newblock Comparison-based learning with rank nets.
\newblock {\em arXiv preprint arXiv:1206.4674}, 2012.

\bibitem{ng2000algorithms}
A.~Y. Ng, S.~J. Russell, et~al.
\newblock Algorithms for inverse reinforcement learning.
\newblock In {\em Icml}, pages 663--670, 2000.

\bibitem{raghavan2006active}
H.~Raghavan, O.~Madani, and R.~Jones.
\newblock Active learning with feedback on features and instances.
\newblock {\em Journal of Machine Learning Research}, 7(Aug):1655--1686, 2006.

\bibitem{ratliff2006maximum}
N.~D. Ratliff, J.~A. Bagnell, and M.~A. Zinkevich.
\newblock Maximum margin planning.
\newblock In {\em Proceedings of the 23rd international conference on Machine
  learning}, pages 729--736. ACM, 2006.

\bibitem{settles2010active}
B.~Settles.
\newblock Active learning literature survey.
\newblock {\em University of Wisconsin, Madison}, 52(55-66):11, 2010.

\bibitem{ziebart2008maximum}
B.~D. Ziebart, A.~L. Maas, J.~A. Bagnell, and A.~K. Dey.
\newblock Maximum entropy inverse reinforcement learning.
\newblock In {\em AAAI}, volume~8, pages 1433--1438. Chicago, IL, USA, 2008.

\bibitem{Ziebart:2008:NLC:1409635.1409678}
B.~D. Ziebart, A.~L. Maas, A.~K. Dey, and J.~A. Bagnell.
\newblock Navigate like a cabbie: Probabilistic reasoning from observed
  context-aware behavior.
\newblock In {\em Proceedings of the 10th International Conference on
  Ubiquitous Computing}, UbiComp '08, pages 322--331, New York, NY, USA, 2008.
  ACM.

\end{thebibliography}
